\documentclass[runningheads]{llncs}

\usepackage{enumitem}
\usepackage{color}
\usepackage{amsmath}
\usepackage{amssymb}
\usepackage{subcaption}
\usepackage{graphicx}
\usepackage{tabularx}

\usepackage{comment}

\usepackage{float}

\begin{document}

\title{H-FCBFormer: Hierarchical Fully Convolutional Branch Transformer for Occlusal Contact Segmentation with Articulating Paper}

\titlerunning{H-FCBFormer}

\author{Ryan Banks\inst{1}\orcidID{0009-0008-6504-7084} \and
Bernat Rovira-Lastra\inst{2,3}\orcidID{0000-0002-4799-6701} \and
Jordi Martinez-Gomis\inst{2}\orcidID{0000-0003-3497-093X} \and
Akhilanand Chaurasia\inst{4}\orcidID{0000-0002-8356-9512} \and
Yunpeng Li\inst{1}\orcidID{0000-0003-4798-541X}
}

\authorrunning{R. Banks et al.}

\institute{University of Surrey, Guildford, Surrey, United Kindgom\\
\email{\{rb01243,yunpeng.li\}@surrey.ac.uk} \and
University of Barcelona, Barcelona, Catalonia, Spain\\
\email{\{brovira,jmartinezgomis\}@ub.edu} \and
IDIBELL, L’Hospitalet de Llobregat, Barcelona, Catalonia, Spain \and 
King George's Medical University, Lucknow, Uttar Pradesh, India\\
\email{chaurasiaakhilanand49@gmail.com}
}

\maketitle

\begin{abstract}
Occlusal contacts are the locations at which the occluding surfaces of the maxilla and the mandible posterior teeth meet. Occlusal contact detection is a vital tool for restoring the loss of masticatory function and is a mandatory assessment in the field of dentistry, with particular importance in prosthodontics and restorative dentistry. The most common method for occlusal contact detection is articulating paper. However, this method can indicate significant medically false positive and medically false negative contact areas, leaving the identification of true occlusal indications to clinicians. To address this, we propose a multiclass Vision Transformer and Fully Convolutional Network ensemble semantic segmentation model with a combination hierarchical loss function, which we name as Hierarchical Fully Convolutional Branch Transformer (H-FCBFormer). We also propose a method of generating medically true positive semantic segmentation masks derived from expert annotated articulating paper masks and gold standard masks. The proposed model outperforms other machine learning methods evaluated at detecting medically true positive contacts and performs better than dentists in terms of accurately identifying object-wise occlusal contact areas while taking significantly less time to identify them. Code is available at \url{https://github.com/Banksylel/H-FCBFormer}.

\end{abstract}

\section{Introduction}

Occlusal Contact Detection (OCD) and analysis play a pivotal role in discerning contact locations between mandibular and maxillary teeth. OCD methods are used extensively in the fields of restorative dentistry, prosthodontics, orthodontics, and implantology, contributing significantly to the diagnosis, treatment, and prognosis of certain dental treatments \cite{cite109} and is an important indicator of the masticatory system’s functional balance. Although several computerised systems, including the T-Scan system, are available for occlusal assessment \cite{cite201}, articulating paper is the most used in clinical practice as it allows the rapid location of occlusal contacts \cite{cite202}, and silicone occlusal registration offers the highest accuracy in a research setting \cite{cite203}. However, indicated contacts for some OCD methods can be ambiguous and incorrect in terms of the size and location of indications compared to the true occlusal contact area, leading to false positive or false negative occlusal contacts, which must be detected intraoperatively by the expertise of clinicians \cite{cite104}.

Deep learning has shown to be a useful tool in the field of dentistry and has found applications from endodontics \cite{cite124} to caries detection in radiographs \cite{cite125}. There has been a rise of medical imaging specific models for semantic segmentation, including U-Net \cite{cite116} and Fully Convolutional Branch transFormer (FCBFormer) \cite{cite13}, where each pixel in an image is predicted as a class or `Background'. However, to the best of our knowledge, there has been no prior work on the application of deep learning to detect occlusal contacts, in spite of the unmet need in the automatic identification of occlusal contacts. Challenges include the development of effective deep learning models and training strategy with the limited availability of gold standard segmentation masks. 

\begin{figure}[htbp]
    \begin{subfigure}{0.24\columnwidth}
        \centering \includegraphics[width=0.99\linewidth]{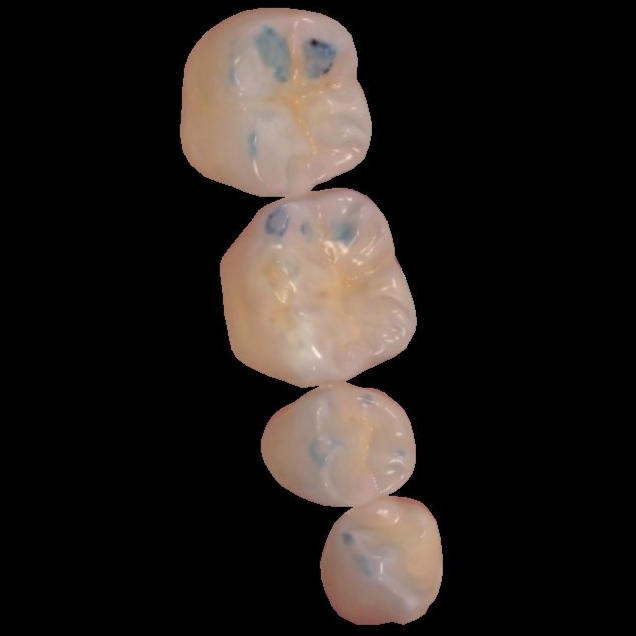}
        \caption{AP Image}
        \label{fig:img103image}
    \end{subfigure}
    \hfill
    \begin{subfigure}{0.24\columnwidth}
        \centering \includegraphics[width=\linewidth]{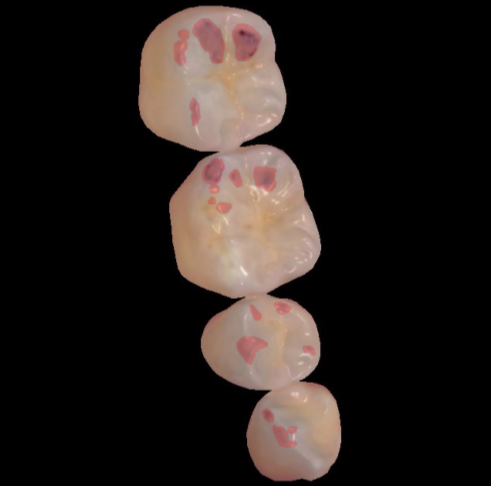}
        \caption{AP Mask}
        \label{fig:img103AP}
    \end{subfigure}
    \hfill
    \begin{subfigure}{0.24\columnwidth}
        \centering \includegraphics[width=0.99\linewidth]{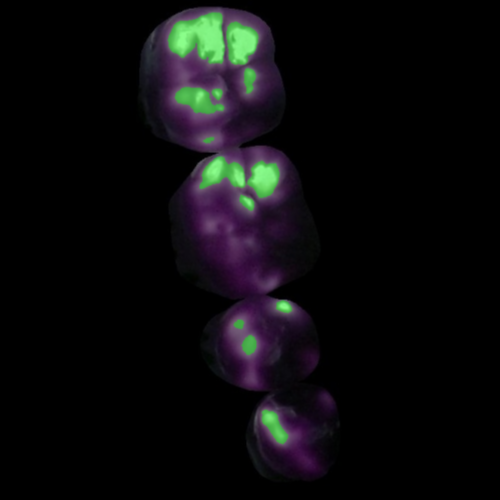}
        \caption{OFR Mask}
        \label{fig:img103OFR}
    \end{subfigure}
    \hfill
    \begin{subfigure}{0.24\columnwidth}
        \centering \includegraphics[width=0.989\linewidth]{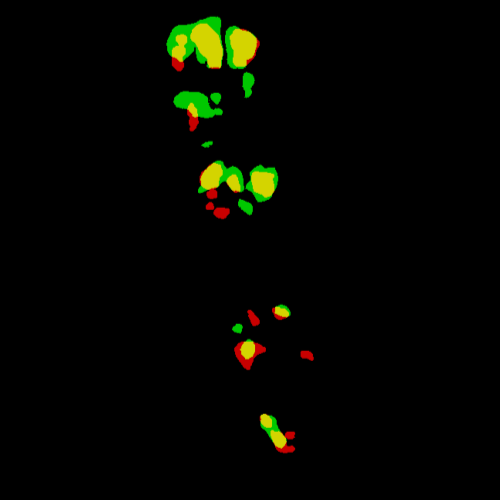}
        \caption{Agreed Location}
        \label{fig:img103Agreed}
    \end{subfigure}

    \caption{Images containing various OCI methods' digital images and masks. Images contain: (a), AP $100\mu m$ Passive AP image with no mask overlay; (b), AP $100\mu m$ Passive image with corresponding FULL contact AP mask in red; (c), OFR image with corresponding OFR $200\mu m$ contact mask in green; (d), no image with AP mask (b) in red, OFR mask (c) in green, and intersection/agreed location (MTP mask) of both (b) and (c) in yellow.}
    \label{fig:img103}
\end{figure}

There have been a number of studies evaluating the viability of different OCD methods \cite{cite128,cite129}. Specifically, Rovira-Lastra et al. \cite{cite1} collected images of five distinct OCD methods on teeth with varying method sensitivities/thickness taken for the right mandibular posterior teeth of 32 patients. Expert annotated binary masks were also collected for each OCD method, which highlights the object-wise indications by the method that indicate a true occlusal contact location, the quality of which was subject to agreement by two expert annotators. The OCD methods evaluated consist of Articulating Paper (AP), Occlufast CAD (OFC), Virtual Occlusion of Intra-Oral Scan (IOS), T-Scan, and the gold standard method Occlufast Rock (OFR) i.e. bite silicone occlusal registration. Examples of an AP image and the associated AP and gold standard OFR masks are shown in Fig. \ref{fig:img103}.

The task we are addressing is to automatically detect the true occlusal contact locations within images of the right mandibular posterior teeth, containing AP ink indications on the teeth. We approach this by detecting two types of occlusal contact masks. Firstly, we aim to detect `object-wise' FULL contacts, which are AP indications that partially indicate an occlusal contact, depicted as red area in Fig. \ref{fig:img103AP}. Secondly, we aim to detect Medically True Positive (MTP) occlusal contacts, which are `pixel-wise' contacts that are contained within the boundaries of the AP ink indications, represented as yellow area in Fig. \ref{fig:img103Agreed}. We define pixel-wise contacts as indicated locations that fully indicate the correct occlusal contact area to a high degree of accuracy, OFR 200$\mu m$ seen in Fig. \ref{fig:img103OFR} indicates pixel-wise contacts. We chose to exclusively use AP images for this problem, as they are the most commonly used OCD method in clinical practice, while the AP mask also shows significant disagreement with the gold standard OFR mask, according to Rovira-Lastra et al. \cite{cite1}.

Our contributions are three-fold: (i) We introduce a method for generating and detecting estimated MTP and Medically False Positive (MFP) multiclass masks, derived from FULL contact AP masks and gold standard OFR 200$\mu m$ masks; (ii) We develop a multiclass adaptation of FCBFormer \cite{cite13} incorporating both a Hierarchical Cross Entropy Loss (HCEL) \cite{cite118} and a proposed Hierarchical Dice Loss (HDL); (iii) We demonstrate that the proposed model outperforms baseline models for MTP, MFP, and FULL object-wise true contact detection, while also outperforming 
dental trained independent observers.

\section{Related Work}

\subsection{Dataset}

In this paper, we utilise the dataset collected by Rovira-Lastra et al. \cite{cite1}, focusing specifically on the use of AP methods for training and evaluation, in contrast to the broader scope of various OCD methods within the dataset. The AP dataset contains 32 patients subjected to four distinct AP methods, of paper thickness $12\mu m$, $40\mu m$, $100\mu m$, and $200\mu m$, respectively. AP films were applied in both active and passive applications, along with test and retest sessions, giving us a total of 512 image-mask pairs utilised for training and validation. The passive application of the film involves the patient biting down three times on the film to imprint ink onto the teeth, followed by the careful removal of the paper after the third bite. Conversely, the active application involves the dentist extracting the paper from the teeth while maintaining the third bite. While the gold standard OFR 200µm image and mask pairs are not directly used in training, an auxiliary set of 70 spatially calibrated OFR 200µm masks is used for generating the proposed MTP masks.

\subsection{FCBFormer}

FCN-Transformer Feature Fusion for Polyp Segmentation (FCBFormer) \cite{cite13} presents an approach to binary semantic segmentation tailored for the Kvasir polyp segmentation dataset \cite{cite115} and the CVC-ClinicDB dataset \cite{cite204}. This work introduces an ensemble architecture comprising two parallel branches of the same input image: a U-Net \cite{cite116} inspired Fully Convolutional Network, and a Pyramid Vision Transformer \cite{cite117}. The resulting feature maps from both branches are concatenated and subsequently fed into the predictor head, to facilitate the reduction of feature maps into a unified binary mask output. This design choice of employing parallel models serves to retain both local and global features, with the Transformer Branch focusing on local features and the Fully Convolutional Branch capturing global context.

\subsection{Hierarchical Loss Function}

A Hierarchical Loss for Semantic Segmentation \cite{cite118} proposes a Hierarchical Cross Entropy Loss function which structures classes into a hierarchy of leaf nodes (detected classes) and parent-classes (sum of child leaf nodes). The loss value is calculated by computing the cross entropy loss \cite{cite119} on nodes in each level of the hierarchy, $\mathcal{L}_{HCEL_{Lev}}$, from the bottom of the hierarchy to the top, where the parent-class is the sum of the child-classes leaf nodes, as seen in Equation \eqref{HCELfunct2}.

\begin{equation}
\begin{split}
	\mathcal{L}_{HCEL_{Lev}} =  -\sum^{N}_{i}\sum^{P}_{p}{y_{ip} log \sum_{k \in L}{\hat{y}_{ipk}}} = -\sum^{N}_{i}\sum^{P}_{p}{y_{ip}log \frac{\sum_{k \in L} exp(x_{ipk})}{\sum^{N}_{j} exp(x_{jp})}} \,\,, 
	\label{HCELfunct2}
\end{split}
\end{equation}
\noindent where $N$ is the number of classes, $P$ is the set of pixels in the image, $y_{ip}$ is the one hot encoded target class $i$ and pixel $p$, $\hat{y}_{ipk}$ is the predicted probabilities for the current level $k$ of the hierarchy, $L$ is the set of leaf nodes levels that contribute to the parent-class, $x_{ipk}$ and $x_{jp}$ are the predicted logit.

The total loss, $\mathcal{L}_{HCEL}$, is computed by summing over all levels of the hierarchy, as shown in Equation \eqref{HCELfunct3}.

\begin{equation}
\begin{split}
	\mathcal{L}_{HCEL} = \sum^{Nlev}_{l}{\mathcal{L}_{HCEL_{Levl}}} \,\,,
	\label{HCELfunct3}
\end{split}
\end{equation}
\noindent where $Nlev$ is the number of levels in the hierarchy.

\section{Materials and Method}

\subsection{H-FCBFormer}

\subsubsection{Model Design}

We propose to adapt the binary semantic segmentation model FCBFormer \cite{cite13} for a multiclass data setting. To achieve this, we add additional output nodes to the model based on the number of classes in the relevant dataset while also changing the loss function and metrics to a multiclass implementation. 

\begin{figure}[!ht]
    \centering
    \includegraphics[width=0.95\linewidth]{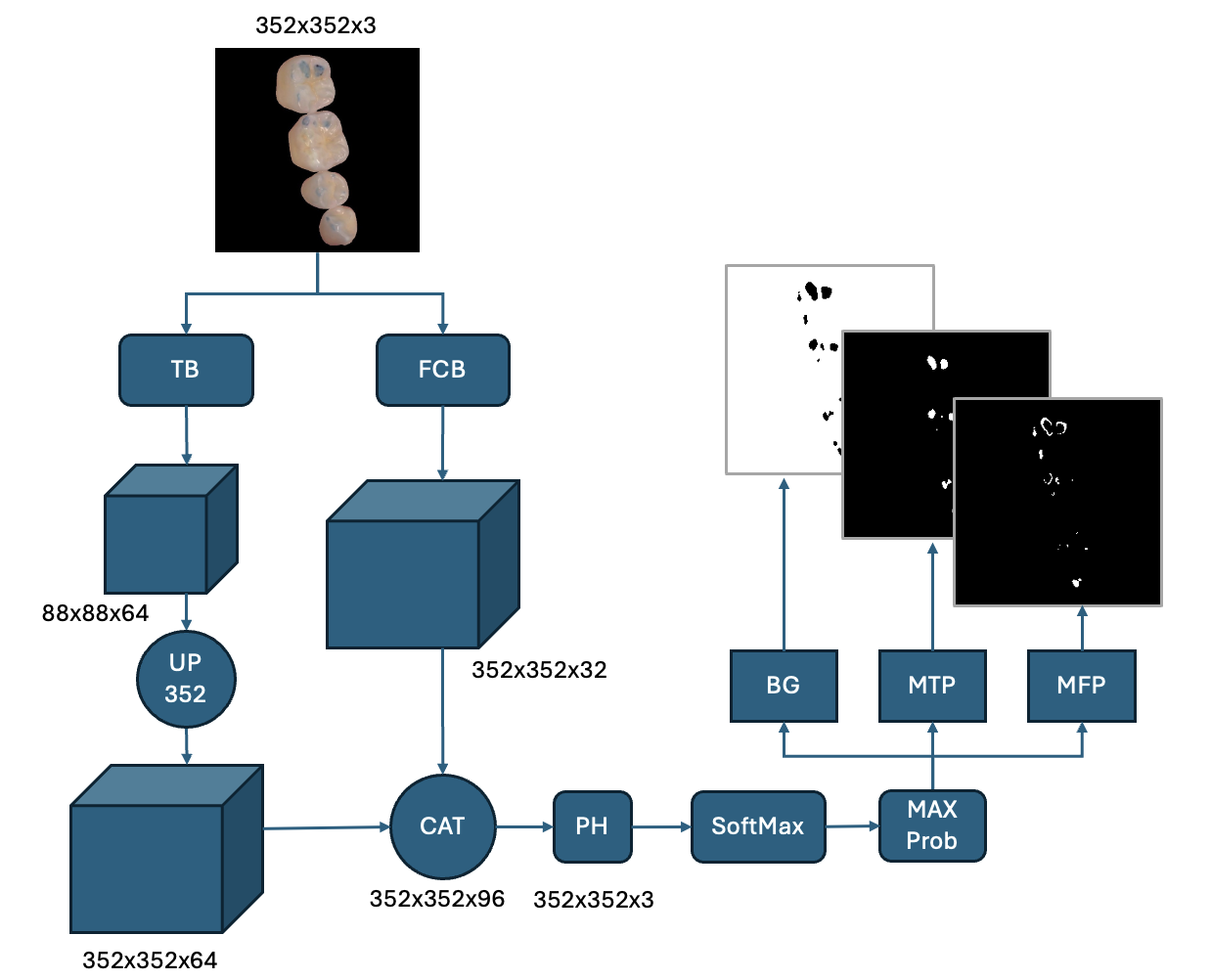}
    \caption{Diagram of the H-FCBFormer multiclass architecture in the inference stage.}
    \label{fig:diagram1}
\end{figure}

The structure of H-FCBFormer in the inference stage, is shown in Fig. \ref{fig:diagram1}. A $352\times352\times3$ RGB image is passed into both the Transformer Branch (TB) and the Fully Convolutional Branch (FCB), which outputs a $88\times88\times64$ and $352\times352\times32$ feature map, respectively. The TB $88\times88\times64$ feature map is then upsampled (UP) to $352\times352\times64$ before being concatenated (CAT) with the FCB $352\times352\times32$ feature map. The concatenated feature maps are then passed into the Predictor Head (PH) with softmax activation to produce a $352\times352\times3$ probability mask of classes Background (BG), Medically True Positive (MTP) and Medically False Positive (MFP). A positive 255 value pixel is assigned to the highest probability class for a given pixel in the image, where the sum of the probabilities of all classes is 1.

\subsubsection{Loss Function}

Our approach leverages the hierarchical loss framework introduced by Muller et al. \cite{cite118}, using the same architecture to compute a hierarchy-aware loss in conjunction with a combination cross entropy and a dice loss function. Specifically, the proposed approach uses the summation of the Hierarchical Cross-Entropy Loss (HCEL) by Muller et al. \cite{cite118} and the loss function we propose, Hierarchical Dice Loss (HDL). We propose this combination hierarchical loss function as the dice loss excels with segmentation performance, but it is susceptible to adversarial attacks and may exhibit poor generalisation, contrasting with the opposite characteristics of the cross-entropy loss \cite{cite120}.

\begin{figure}[!ht]
    \centering
    \includegraphics[width=0.8\linewidth]{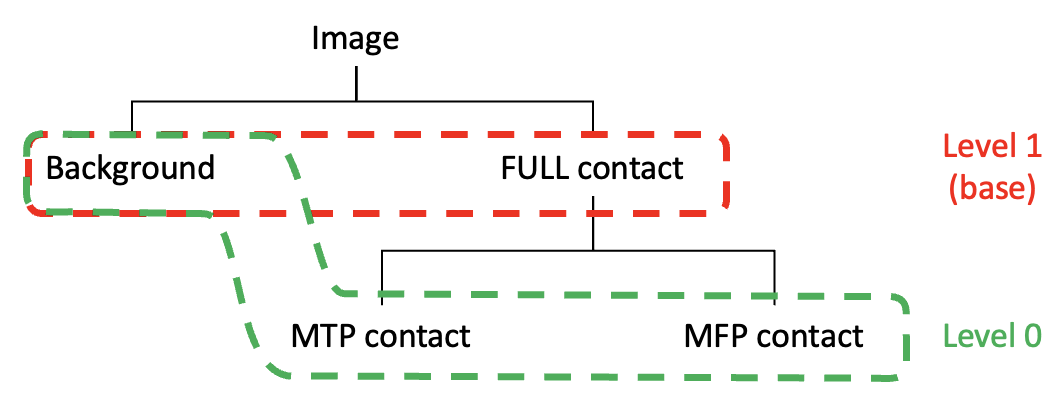}
    \caption{Image showing the class hierarchy of the occlusal contact dataset. Image is not considered a class in the loss calculation.}
    \label{fig:img202}
\end{figure}

We structure the dataset hierarchically, with `Background', `MTP contacts', and `MFP contacts' as primary leaf node classes, and a parent-class of MTP and MFP classes denoted as `FULL contact', as depicted in Fig. \ref{fig:img202}. Although the primary focus of our analysis lies on the detection of MTP and FULL contact classes, the identification of MFP contacts is necessary for discerning FULL contact locations, as it is comprised of the union of MTP and MFP locations.

Referring to Fig. \ref{fig:img202}, the computation of our hierarchical loss function entails initially calculating the level loss value of all leaf nodes (Background, MTP contact, MFP contact). Followed by calculating the level loss value of the parent-classes (Background, FULL contact) of one level higher from the lowest level within the hierarchy, by summing the output probabilities of the previous child-nodes for each of the new parent-class. The level loss value is calculated for each level of the hierarchy until the base classes (Background, FULL contact) are reached. Leaf nodes in upper hierarchy levels (Background) contribute to all level loss calculations in lower levels than itself. The final loss value is calculated by summing the level loss values of all levels in the hierarchy.

\textbf{Hierarchical Dice Loss} is a loss function proposed in this paper, designed in accordance with the methodology of HCEL. This approach involves the computation of multiclass Dice Loss \cite{cite122} at every hierarchical level, as depicted in Equation \eqref{HDLfunct2}. The aggregate HDL is obtained by summing the loss values across all hierarchical levels, as expressed in Equation \eqref{HDLfunct3}.

\begin{equation}
\begin{split}
	\mathcal{L}_{HDL_{lev}} = 1 - 2 \frac{\sum^{N}_{i} \sum^{P}_{p}{y_{ip} \sum_{k \in L}{ \hat{y}_{ipk}}} + \epsilon}{\sum^{N}_{i} \sum^{P}_{p}({y_{ip}} + \sum_{k \in L}{ \hat{y}_{ipk}} ) + \epsilon} \\
    = 1 - 2 \frac{\sum^{N}_{i} \sum^{P}_{p}{y_{ip} \frac{\sum_{k \in L} exp(x_{ipk})}{\sum^{N}_{j}exp(x_{jp})}} + \epsilon}{\sum^{N}_{i} \sum^{P}_{p} \left( {y_{ip}} + \frac{\sum_{k \in L} exp(x_{ipk})}{\sum^{N}_{j}exp(x_{jp})} \right)  + \epsilon} \,\,,
	\label{HDLfunct2}
\end{split}
\end{equation}
\begin{equation}
\begin{split}
	\mathcal{L}_{HDL} = \sum^{Nlev}_{l}{\mathcal{L}_{HDL_{Lev l}}} \,\,,
	\label{HDLfunct3}
\end{split}
\end{equation}

\noindent where $\epsilon$ is a small constant to prevent division by 0, a practice we inherit from \cite{cite123}. $Nlev$ is 2 for our dataset.

Following the computation of HCEL as proposed by Muller et al. \cite{cite118}, and our proposed HDL, the ultimate loss value is derived by summing both HCEL and HDL, as expressed in Equation \eqref{fullLossFunct}.

\begin{equation}
\begin{split}
	\mathcal{L} = \mathcal{L}_{HCEL} + \mathcal{L}_{HDL} \,\,.
	\label{fullLossFunct}
\end{split}
\end{equation}

\subsection{Creating Medically True and False Positive Masks}

In the absence of pre-existing MTP masks, we produce their approximations based on the methodology outlined in the work by Rovira-Lastra et al.\cite{cite1}. Utilising OFR $200\mu m$ gold standard masks directly as targets for AP images can yield reasonably accurate masks, reflecting the genuine pixel-wise contact area to a high degree of accuracy. However, areas present in the OFR mask that are missed by the AP method lacks physical indication (AP ink) in the image. This scenario could lead to suboptimal performance with a segmentation model and pose challenges in distinguishing inked and non-inked contact locations. Additionally, OFR masks have a reproducibility agreement ranging from 0.92 to 0.98, indicating slight errors in identifying the contact area for the same patient. Therefore, we consider that a more effective indication of pixel-wise contact locations would be the overlapping region between test and retest OFR $200\mu m$ masks.

\begin{figure}[htbp]
    \begin{subfigure}{0.25\columnwidth}
        \centering \includegraphics[width=0.85\linewidth]{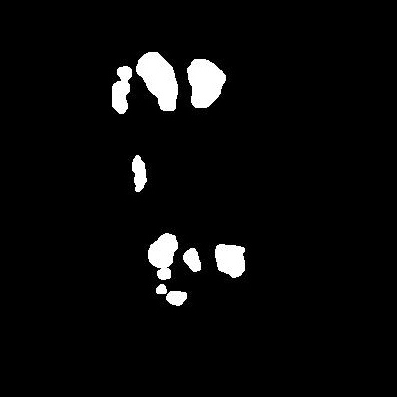}
        \caption{AP Mask}
    \end{subfigure}
    \hfill
    \begin{subfigure}{0.25\columnwidth}
        \centering \includegraphics[width=0.85\linewidth]{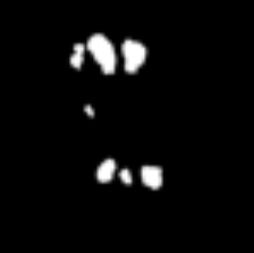}
        \caption{OFR Td$\cap$Rd}
    \end{subfigure}
    \hfill
    \begin{subfigure}{0.25\columnwidth}
        \centering \includegraphics[width=0.85\linewidth]{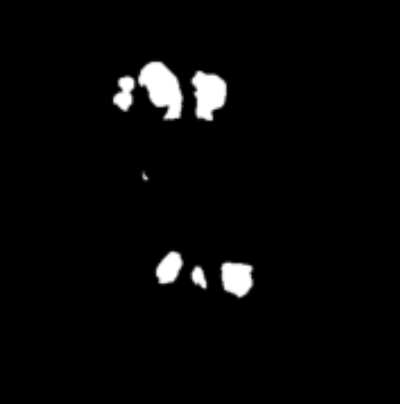}
        \caption{Intersection}
    \end{subfigure}
    \medskip
    
    \begin{subfigure}{0.25\columnwidth}
        \centering \includegraphics[width=0.85\linewidth]{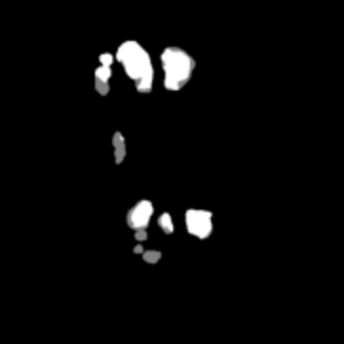}
        \caption{MTP and MFP}
    \end{subfigure}
    \hfill
    \begin{subfigure}{0.25\columnwidth}
        \centering \includegraphics[width=0.85\linewidth]{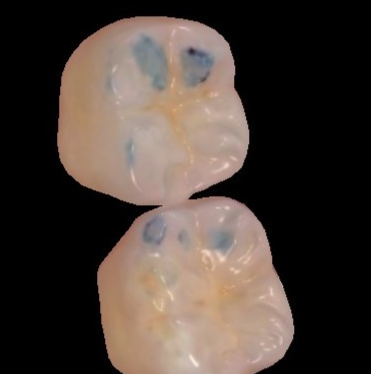}
        \caption{Image}
    \end{subfigure}
    \hfill
    \begin{subfigure}{0.25\columnwidth}
        \centering \includegraphics[width=0.85\linewidth]{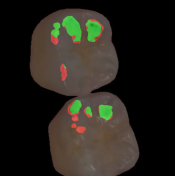}
        \caption{Image and Mask}
    \end{subfigure}
    \caption{Images containing masks for each stage of the mask generation process, for patient 01 retest AP $100\mu m$ Passive. (a) Articulating Paper mask, (b) intersection of the OFR $200\mu m$ test (Td) and retest (Rd) mask, (c) intersection of (a) and (b), (d) final MTP (white) and MFP (grey) multiclass mask, (e) digital image, and (f) is image (e) with an overlaid green/red version of (d) where red is MFP contacts and green is MTP contacts.}
    \label{fig:img107}
\end{figure}

We propose a method of generating estimated Medically True Positive (MTP) masks, that indicate the pixel-wise true occlusal contacts, visualised in Fig. \ref{fig:img107}. We use the intersection between the OFR $200\mu m$ test and retest masks to find the absolute true contact area. We then compute the intersection between the OFR $200\mu m$ test/retest intersection mask and the AP masks for each $\mu m$ thickness of the same patient. The resulting mask indicates the MTP pixel-wise true occlusal contacts within the boundaries of the AP ink. This is considered the mask for Class 1: True Postive Contacts (MTP contacts). We use the MTP contact mask to create a masks indicating the MFP contacts, by removing the intersection of the MTP masks from the original AP mask. This is considered as the mask for Class 2: False Positive Contacts (MFP contacts). Each positive pixel within these class maps is assigned a unique pixel value within a combined grayscale multiclass mask, with non-positive class pixels categorised under the Background Class 0. Although FULL contact masks are not explicitly detected as a distinct class, they can be assessed as the union of the MTP and MFP contact masks.

\subsection{Data Acquisition and Pre-Processing}

The dataset is comprised of AP images from 32 patients (512 image-mask pairs), divided into four folds of alternating train/validation sets of 24 patients (384 image-mask pairs) and 8 patients (128 image-mask pairs), respectively. A 4-fold cross-validation approach is employed, where four separate models are trained with identical hyperparameters.

The original images in the dataset possessed a default resolution of $3840 \times 2160$, containing significant areas of empty space. Automatic resizing of these images to dimensions of $352 \times 352$ by the model would lead to distortion and notable decline in the quality of the predicted masks. To mitigate this issue, we opt to resize and reposition both the images and masks to a resolution of $1000 \times 1000$, encompassing the relevant area of the images. Additionally, we ensure the retention of spatial calibration by applying the same transformation to all images from the same patient.

\subsection{Implementation Details}

We utilise pre-trained backbone weights from \cite{cite13} for both the Transformer Branch and the Fully Convolutional Branch to fine tune the model to our small data regime dataset. However, we exclude the Predictor Head from using pre-trained weights due to its shallow architecture and different number of output nodes. We fine-tune H-FCBFormer with our data to predict RGB images resized to $352 \times 352$ pixels, with a target grayscale multiclass semantic segmentation map of 3 classes (Background, MTP contact, MFP contact). We use a PyTorch implementation of this model. Random augmentation is used for the training input and target pairs as they are loaded at each epoch to supplement the number of data samples. We run experiments at 400 epochs with a batch size of 5, saving the weights at the epoch with the best validation Dice score and last epoch, evaluating using the best weights. We use the AdamW optimiser, at a reduced learning rate of $10^{-4}$ that reduces by a factor of 2 when the Dice performance does not improve over 3 epochs until it reaches $10^{-6}$. A workstation with an NVIDIA RTX 3090 GPU and 64GB of RAM is used for training.

All baseline model implementations use the same augmentation, image input size, evaluation, classes and data splits. Each model uses pre-trained weights unique to each model before re-training begins on our dataset. Each model used their respective loss functions and types of hyperparameters as stated in their respective papers, except for FCBFormer which has its loss function and output nodes adapted to a multiclass data setting. The hyperparameters of all methods are manually fine tuned over 10 training loops before they are evaluated.

\section{Results and Discussion}

\subsection{Quantitative Results}

\subsubsection{Segmentation Metrics and Evaluation} 
Semantic segmentation metrics are used to evaluate the performance of each model at predicting the object-wise FULL contact area, the estimated pixel-wise MTP contact area and the MFP contact area. Metrics include IoU = $\frac{TP}{TP+FP+FN}$, Precision = $\frac{TP}{TP+FP}$, Recall =  $\frac{TP}{TP+FN}$, Dice = $\frac{2TP}{2TP+FP+FN}$. True Positive (TP), True Negative (TN), False Positive (FP), and False Negative (FN) are the numbers of pixels in an image for a specific class. Final metrics are presented as the mean (standard deviation) metrics for all validation images present within the four folds, where the standard deviation is calculated over all validation images and averaged over all folds.

Table \ref{tab:table111} presents the segmentation metrics of the proposed model H-FCBFormer, compared with two baseline models multiclass FCBFormer and multiclass U-Net. The proposed H-FCBFormer model exhibits superior performance across most metrics compared to the baseline models, specifically for IoU, Dice, and Precision. Both the proposed model and U-Net exhibit higher Precision over Recall, prioritising reduced false positives at the expense of increased false negatives. Despite that the FULL contact is not designated as a distinct class, all models have significantly improved performance for this category compared to the other classes, characterised by higher metrics and lower average standard deviation. Additionally, the proposed model outperforms the other models for FULL contact segmentation, potentially attributable to the hierarchy-aware architecture of the loss function.

\begin{table}[hbtp]
\caption{Table comparing the mean segmentation metrics and (mean standard deviation) of various models for relevant classes, MTP, MFP and FULL contact detection tasks.}
\label{tab:table111}
\centering
\resizebox{\columnwidth}{!}{%
\begin{tabular}{ll|c|c|c|c|}
\cline{3-6}
\multicolumn{2}{l|}{}                                                                            & \textbf{IoU}                                                                         & \textbf{Dice}                                                                        & \textbf{Precision}                                                                   & \textbf{Recall}                                                                      \\ \cline{1-6}
\multicolumn{2}{|l|}{\textbf{\begin{tabular}[c]{@{}l@{}}FCBFormer\end{tabular}}}   &                                                                                      &                                                                                      &                                                                                      &                                                                                      \\

\multicolumn{1}{|l}{} & \begin{tabular}[c]{@{}l@{}}MTP\end{tabular}         & \begin{tabular}[c]{@{}c@{}}0.400 ($\pm$ 0.126)\end{tabular}          & \begin{tabular}[c]{@{}c@{}}0.577 ($\pm$ 0.141)\end{tabular}          & \begin{tabular}[c]{@{}c@{}}0.536 ($\pm$ 0.159)\end{tabular}          & \begin{tabular}[c]{@{}c@{}}\textbf{0.628} ($\pm$ 0.176)\end{tabular} \\
\multicolumn{1}{|l}{} & \begin{tabular}[c]{@{}l@{}}MFP\end{tabular}                                                               & \begin{tabular}[c]{@{}c@{}}0.258 ($\pm$ 0.084)\end{tabular}          & \begin{tabular}[c]{@{}c@{}}0.403 ($\pm$ 0.108)\end{tabular}          & \begin{tabular}[c]{@{}c@{}}0.351 ($\pm$ 0.120)\end{tabular}          & \begin{tabular}[c]{@{}c@{}}\textbf{0.518} ($\pm$ 0.131)\end{tabular} \\
\multicolumn{1}{|l}{} & \begin{tabular}[c]{@{}l@{}}FULL\end{tabular} & \begin{tabular}[c]{@{}c@{}}0.334 ($\pm$ 0.078)\end{tabular}          & \begin{tabular}[c]{@{}c@{}}0.420 ($\pm$ 0.080)\end{tabular}          & \begin{tabular}[c]{@{}c@{}}0.564 ($\pm$ 0.175)\end{tabular}          & \begin{tabular}[c]{@{}c@{}}0.449 ($\pm$ 0.080)\end{tabular}           \\
\multicolumn{2}{|l|}{\textbf{U-Net}}                                                              &                                                                                      &                                                                                      &                                                                                      &                                                                                      \\
\multicolumn{1}{|l}{} & \begin{tabular}[c]{@{}l@{}}MTP\end{tabular}         & \begin{tabular}[c]{@{}c@{}}0.369 ($\pm$ 0.143)\end{tabular}          & \begin{tabular}[c]{@{}c@{}}0.555 ($\pm$ 0.165)\end{tabular}          & \begin{tabular}[c]{@{}c@{}}0.628 ($\pm$ 0.174)\end{tabular}          & \begin{tabular}[c]{@{}c@{}}0.489 ($\pm$ 0.188)\end{tabular}          \\
\multicolumn{1}{|l}{} & \begin{tabular}[c]{@{}l@{}}MFP\end{tabular}                                                              & \begin{tabular}[c]{@{}c@{}}0.256 ($\pm$ 0.088)\end{tabular}          & \begin{tabular}[c]{@{}c@{}}0.399 ($\pm$ 0.012)\end{tabular}          & \begin{tabular}[c]{@{}c@{}}0.394 ($\pm$ 0.502)\end{tabular}          & \begin{tabular}[c]{@{}c@{}}0.438 ($\pm$ 0.146)\end{tabular}          \\
\multicolumn{1}{|l}{} & \begin{tabular}[c]{@{}l@{}}FULL\end{tabular} & \begin{tabular}[c]{@{}c@{}}0.524 ($\pm$ 0.184)\end{tabular}          & \begin{tabular}[c]{@{}c@{}}0.666 ($\pm$ 0.180)\end{tabular}           & \begin{tabular}[c]{@{}c@{}}0.907 ($\pm$ 0.073)\end{tabular}          & \begin{tabular}[c]{@{}c@{}}0.449 ($\pm$ 0.193)\end{tabular}          \\ \cline{1-6}
\multicolumn{2}{|l|}{\textbf{H-FCBFormer}}                                                              &                                                                                      &                                                                                      &                                                                                      &                                                                                      \\
\multicolumn{1}{|l}{} & \begin{tabular}[c]{@{}l@{}}MTP\end{tabular}         & \begin{tabular}[c]{@{}c@{}}\textbf{0.427} ($\pm$ 0.128)\end{tabular} & \begin{tabular}[c]{@{}c@{}}\textbf{0.585} ($\pm$ 0.139)\end{tabular} & \begin{tabular}[c]{@{}c@{}}\textbf{0.643} ($\pm$ 0.161)\end{tabular} & \begin{tabular}[c]{@{}c@{}}0.571 ($\pm$ 0.156)\end{tabular}          \\
\multicolumn{1}{|l}{} & \begin{tabular}[c]{@{}l@{}}MFP\end{tabular}                                                              & \begin{tabular}[c]{@{}c@{}}\textbf{0.302} ($\pm$ 0.091)\end{tabular} & \begin{tabular}[c]{@{}c@{}}\textbf{0.456} ($\pm$ 0.113)\end{tabular} & \begin{tabular}[c]{@{}c@{}}\textbf{0.510} ($\pm$ 0.147)\end{tabular} & \begin{tabular}[c]{@{}c@{}}0.447 ($\pm$ 0.136)\end{tabular}          \\
\multicolumn{1}{|l}{} & \begin{tabular}[c]{@{}l@{}}FULL\end{tabular} & \begin{tabular}[c]{@{}c@{}}\textbf{0.662} ($\pm$ 0.118)\end{tabular} & \begin{tabular}[c]{@{}c@{}}\textbf{0.789} ($\pm$ 0.095)\end{tabular} & \begin{tabular}[c]{@{}c@{}}\textbf{0.908} ($\pm$ 0.064)\end{tabular} & \begin{tabular}[c]{@{}c@{}}\textbf{0.713} ($\pm$ 0.131)\end{tabular} \\ \cline{1-6}
\end{tabular}
}
\end{table}

\subsubsection{Independent Observer Study} 
Within this study, we compare the performance of 4 individual independent observers at creating object-wise FULL contact masks, against our model's performance at predicting the same object-wise FULL contact masks. We requested 4 dental trained independent observers to annotate a small 8 patient (128 images) AP subset, randomly sampled from the existing full dataset 32 patient (512 images). Each dentist was tasked with indicate the object-wise FULL contacts under the same instructions and method as the annotators in the study by Rovira-Lastra et al. \cite{cite1}. However, the 4 dentists were asked not to consult with anyone, while also recording the time taken to complete the task. The predictions generated by the proposed model and the masks created by the 4 dentists are evaluated against the object-wise FULL contact target masks from the existing dataset. The target object-wise FULL contact masks from the existing study are considered the ground truth for this independent observer study, as the annotators were allowed to consult with each other with their higher combined experience and additional time spent in verifying the quality of the masks. The aim of this study is to determine if the proposed model's performance and time to predict object-wise FULL contact masks are superior to the individual performance of each of the 4 dentists.

The proposed model exhibits increased performance for IoU, Dice, and Recall, compared to all of the dentists evaluated, seen in Table \ref{tab:table113}. Even though Dentist 1 has increased precision compared to the proposed model, Dentist 1 has a significantly lower recall. Dentists 1, 2, and 3 demonstrate similar and improved performance across most metrics when compared to Dentist 4, suggesting that these dentists created masks of higher and similar qualities. Dentists 1, 2, and 4 exhibit significantly higher Precision than Recall, indicating a tendency to incorrectly annotate AP indications as false contacts compared to the gold standard. Conversely, the proposed model's predictions and Dentist 3 have a more balanced Precision and Recall, implying a more accurate identification of the correct object-wise FULL contact area. Additionally, the proposed method takes negligible time to produce object-wise FULL contact masks compared with any of the four dentists. These findings indicate superior performance of the proposed method in identifying correct object-wise FULL contact locations compared to individual dentists, coupled with its significantly increased speed of inference.

\begin{table}[hbtp]
\caption{Table comparing mean metrics (standard deviation) for the proposed model predictions and the manual annotations of 4 dentists, for a small randomly selected 8 patient subset of AP object-wise FULL contact masks. We also compare the average time taken to create the FULL contact masks per image in seconds, for each dentist and our model.}
\label{tab:table113}
\centering
\resizebox{\columnwidth}{!}{%
\begin{tabular}{c|c|c|c|c|c|}
\cline{2-6}
\multicolumn{1}{l|}{}                                  & \textbf{IoU}                 & \textbf{Dice}                & \textbf{Precision}           & \textbf{Recall}       &\textbf{Time}                       \\ \hline
\multicolumn{1}{|c|}{Dentist 1}                        & 0.427 ($\pm$ 0.110)          & 0.590 ($\pm$ 0.112)          & \textbf{0.902} ($\pm$ 0.080) & 0.451 ($\pm$ 0.119)       & 109.9s          \\
\multicolumn{1}{|c|}{Dentist 2}                        & 0.587 ($\pm$ 0.121)          & 0.731 ($\pm$ 0.107)          & 0.866 ($\pm$ 0.136)          & 0.650 ($\pm$ 0.122)       & 135.4s          \\
\multicolumn{1}{|c|}{Dentist 3}                        & 0.571 ($\pm$ 0.109)          & 0.721 ($\pm$ 0.095)          & 0.778 ($\pm$ 0.134)          & 0.692 ($\pm$ 0.109)       & 136.0s          \\
\multicolumn{1}{|c|}{Dentist 4}                        & 0.247 ($\pm$ 0.098)          & 0.386 ($\pm$ 0.120)          & 0.814 ($\pm$ 0.133)          & 0.265 ($\pm$ 0.110)       & 104.5s           \\ \cline{1-6}
\multicolumn{1}{|c|}{Proposed}          & \textbf{0.697} ($\pm$ 0.094) & \textbf{0.817} ($\pm$ 0.070) & 0.871 ($\pm$ 0.070)          & \textbf{0.781} ($\pm$ 0.107)        & \textbf{0.2s} \\ \hline
\end{tabular}
}
\end{table}

\subsection{Qualitative Results}

We compare the images of two AP images with corresponding predicted and target masks, from each model. The masks are displayed where red denotes the predicted mask (False Positive), green denotes the target mask (False Negative) and yellow denotes the agreed location of green and red (True Positive). Comparisons are made across MTP, MFP, and FULL contact classes for all deep learning models, focusing on patient (a) 01 retest AP $100\mu m$ Passive and patient (b) 01 retest AP $12\mu m$ Active. 

\begin{figure}[!ht]
\begin{minipage}{0.87\linewidth}
\begin{subfigure}{0.49\columnwidth}
    \begin{subfigure}{0.32\columnwidth}
        \centering \includegraphics[width=\linewidth]{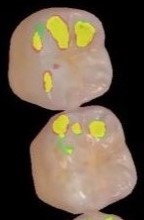}
    \end{subfigure}
    \hfill
    \begin{subfigure}{0.32\columnwidth}
        \centering \includegraphics[width=\linewidth]{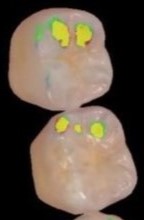}
    \end{subfigure}
    \hfill
    \begin{subfigure}{0.32\columnwidth}
        \centering \includegraphics[width=\linewidth]{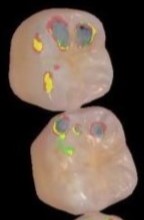}
    \end{subfigure}
\end{subfigure}
\begin{minipage}{0.15\linewidth}
\end{minipage}
\begin{subfigure}{0.49\columnwidth}
    \begin{subfigure}{0.32\columnwidth}
        \centering \includegraphics[width=\linewidth]{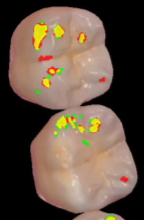}
    \end{subfigure}
    \hfill
    \begin{subfigure}{0.32\columnwidth}
        \centering \includegraphics[width=\linewidth]{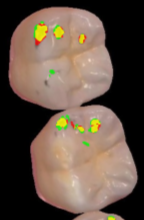}
    \end{subfigure}
    \hfill
    \begin{subfigure}{0.32\columnwidth}
        \centering \includegraphics[width=\linewidth]{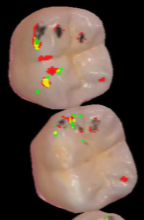}
    \end{subfigure}
\end{subfigure}
\end{minipage}\hfill
    \begin{minipage}{0.10\linewidth}
    \begin{subfigure}{\columnwidth}
    \caption*{FCBFormer}
    \end{subfigure}
    \end{minipage}
    \medskip

\begin{minipage}{0.87\linewidth}
\begin{subfigure}{0.49\columnwidth}
    \begin{subfigure}{0.32\columnwidth}
        \centering \includegraphics[width=\linewidth]{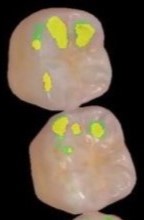}
    \end{subfigure}
    \hfill
    \begin{subfigure}{0.32\columnwidth}
        \centering \includegraphics[width=\linewidth]{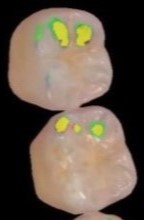}
    \end{subfigure}
    \hfill
    \begin{subfigure}{0.32\columnwidth}
        \centering \includegraphics[width=\linewidth]{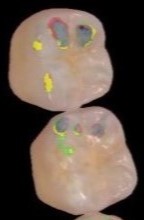}
    \end{subfigure}
\end{subfigure}
\begin{minipage}{0.15\linewidth}
\end{minipage}
\begin{subfigure}{0.49\columnwidth}
    \begin{subfigure}{0.32\columnwidth}
        \centering \includegraphics[width=\linewidth]{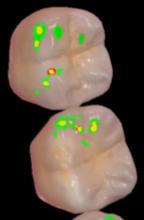}
    \end{subfigure}
    \hfill
    \begin{subfigure}{0.32\columnwidth}
        \centering \includegraphics[width=\linewidth]{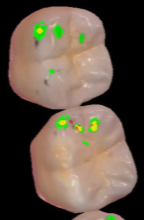}
    \end{subfigure}
    \hfill
    \begin{subfigure}{0.32\columnwidth}
        \centering \includegraphics[width=\linewidth]{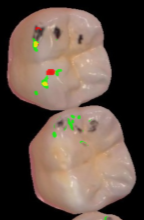}
    \end{subfigure}
\end{subfigure}
\end{minipage}\hfill
    \begin{minipage}{0.10\linewidth}
    \begin{subfigure}{\columnwidth}
    \caption*{U-Net}
    \end{subfigure}
    \end{minipage}
    \medskip

\begin{minipage}{0.87\linewidth}
\begin{subfigure}{0.49\columnwidth}
    \begin{subfigure}{0.32\columnwidth}
        \centering \includegraphics[width=\linewidth]{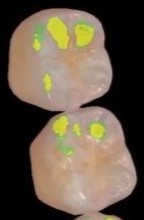}
        \caption*{FULL}
    \end{subfigure}
    \hfill
    \begin{subfigure}{0.32\columnwidth}
        \centering \includegraphics[width=\linewidth]{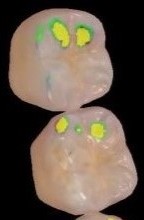}
        \caption*{MTP}
    \end{subfigure}
    \hfill
    \begin{subfigure}{0.32\columnwidth}
        \centering \includegraphics[width=\linewidth]{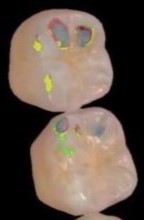}
        \caption*{MFP}
    \end{subfigure}
\caption{$100\mu m$ Passive}
\label{fig:img108a}
\end{subfigure}
\begin{minipage}{0.15\linewidth}
\end{minipage}
\begin{subfigure}{0.49\columnwidth}
    \begin{subfigure}{0.32\columnwidth}
        \centering \includegraphics[width=\linewidth]{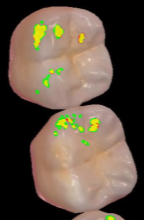}
        \caption*{FULL}
    \end{subfigure}
    \hfill
    \begin{subfigure}{0.32\columnwidth}
        \centering \includegraphics[width=\linewidth]{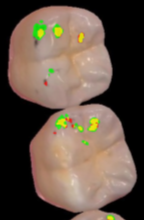}
        \caption*{MTP}
    \end{subfigure}
    \hfill
    \begin{subfigure}{0.32\columnwidth}
        \centering \includegraphics[width=\linewidth]{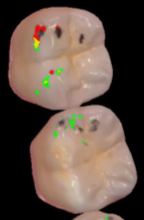}
        \caption*{MFP}
    \end{subfigure}
\caption{$12\mu m$ Active}
\label{fig:img108b}
\end{subfigure}
\end{minipage}\hfill
    \begin{minipage}{0.11\linewidth}
    \begin{subfigure}{\columnwidth}
    \caption*{H-FCB-Former}
    \end{subfigure}
    \end{minipage}

    \caption{Images containing the cropped digital images, predicted masks, and target masks, for patient 01 retest AP $100\mu m$ Passive and patient 01 retest AP $12\mu m$ Active, respectively. Each method is displayed for the FULL contact class, MTP class and MFP class. Green indicates the target mask (FN), red indicates the Predicted mask (FP) and yellow indicates the overlap/agreed location (TP).}
    \label{fig:img108}
\end{figure}

Within Fig. \ref{fig:img108}, subfigures (a) and (b) for the proposed model, show an increased prevalence of FN predictions over FP predictions, as indicated by the expanded green area relative to red. Additionally, the proposed model is shown to predict high-quality masks, characterised as well-defined predictions closely fitting to the objects in the image for both MTP and MFP classes. U-Net also showcases prevalence for FN predictions over FP predictions for the FULL contact class. However, this appears to be achieved through under-predicting all positive predictions for MTP and MFP, as evidenced in subfigures (b) for U-Net. FCBFormer, while proficient in detecting the general locations of contacts, falls short in accurately predicting entire contact areas. This is due to FCBFormer's tendency to over-predict positive predictions, leading to an increased prevalence of FP predictions, as depicted in subfigures (b) for FCBFormer. Across all models, suboptimal performance is observed in detecting the entire contact area and accurately determining the correct class for some small contacts, as evidenced in subfigures (b).

\section{Conclusion and Limitations}

We address the problem of automatically detecting occlusal contact locations, specifically pixel-wise true positive (MTP) contacts and object-wise FULL contacts, from digital images of the widely used yet often inaccurate AP film on teeth. The problem is to identify the correct occlusal contact locations for a patient, by evaluating the accuracy of the AP method itself. This is addressed by determining the MTP and MFP pixel accurate contact locations within the AP ink indications. Given the absence of expert-annotated MTP contact masks, we introduce a method of generating MTP masks from the intersection of AP, OFR 200$\mu m$ test and OFR 200$\mu m$ retest masks. We also propose a hierarchical aware vision transformer and fully convolutional network ensemble model optimised for the nuanced features of medical imaging, with a hierarchical loss function.

The proposed model outperforms 4 independent observers and similar multiclass semantic segmentation models, with our small data regime dataset. The proposed model performs well at indicating object-wise FULL contact and pixel-wise MTP contact masks for images of AP ink on teeth, while taking a less time to predict them. Implying that the proposed method would be useful at improving the accuracy of clinicians when determining the true occlusal contact locations from AP ink on teeth. The increased performance of our model compared to the non-hierarchical version of the model (FCBFormer), is likely attributed to the hierarchical loss function, that adjusts weights with regards to the child-classes individually, and with regards to how the union of all relevant child-classes effects the parent-class. 

As H-FCBFormer performs well on our problem of occlusal contact detection, we also surmise that model will perform well in other medical data domains which have hierarchical data setups, where the the detection of the parent-classes are equally as important as the child-classes they are comprised of. The model architecture could also be adapted for other non-medical hierarchical data settings. However, while our model performs well when trained and evaluated on our proposed method of generating MTP masks, without evaluation to ground truth MTP masks, we cannot accurately assess the quality of the generated masks and model performance in determining high quality MTP contact locations. Additionally, uncertainties persist regarding the model's ability to generalise to unseen samples, attributed to the small sample size (512 images) of the dataset and unbalanced background pixels compared to class pixels. Therefore, the collection of a larger dataset with additional MTP and MFP annotations created by experts, additional experimentation on other hierarchical datasets, and research into weighted hierarchical loss functions, are among the potential directions for future work.

\newpage
\bibliographystyle{splncs04}
\bibliography{strings, refs}
\end{document}